\documentclass{article}
\usepackage{spconf,amsmath,graphicx}
\usepackage[colorlinks=true]{hyperref}

\usepackage{enumitem}
\setlist{nosep, leftmargin=14pt}

\usepackage{mwe} 
\usepackage{multirow}
\usepackage{graphicx}
\usepackage{soul}
\usepackage{amsmath}
\usepackage{amssymb}
\usepackage{multicol}
\usepackage{booktabs}

\usepackage{algorithm, algorithmic}
\usepackage{amsmath, amssymb}
\usepackage[usenames,dvipsnames]{xcolor}

\title{Can Score-based Generative Modeling Effectively Handle Medical Image Classification?}
\name{Sushmita Sarker$^*$, Prithul Sarker$^*$, George Bebis, and Alireza Tavakkoli\thanks{$^*$ Equal contribution}}
\address{Department of Computer Science and Engineering, University of Nevada, Reno, USA}
\begin{document}
%
\maketitle
\begin{abstract}
The remarkable success of deep learning in recent years has prompted applications in medical image classification and diagnosis tasks. 
While classification models have demonstrated robustness in classifying simpler datasets like MNIST or natural images such as ImageNet, this resilience is not consistently observed in complex medical image datasets where data is more scarce and lacks diversity. 
Moreover, previous findings on natural image datasets have indicated a potential trade-off between data likelihood and classification accuracy. 
In this study, we explore the use of score-based generative models as classifiers for medical images, specifically mammographic images. 
Our findings suggest that our proposed generative classifier model not only achieves superior classification results on CBIS-DDSM, INbreast and Vin-Dr Mammo datasets, but also introduces a novel approach to image classification in a broader context. Our code is publicly available at \url{https://github.com/sushmitasarker/sgc_for_medical_image_classification}

\end{abstract}
\section{Introduction}
\label{sec:intro}

A classification task can be approached in two ways, i.e., posterior approximation (discriminative) or likelihood approximation (generative), each offering distinct solutions~\cite{ng2001discriminative}. Discriminative classifiers, directly estimate posterior probabilities $p(c | x)$ based on given inputs $(x)$ and a set of classes $(c)$. In contrast, generative classifiers model the likelihood of inputs conditioned on specific classes $p(x | c)$, determining the category with the highest likelihood as the final decision~\cite{mackowiak2021generative}.

Discriminative models have traditionally excelled in efficiency and supervised learning, yet they may exploit spurious correlations, limiting reliability on unseen data. Generative classifiers, while generally slower, provide robustness by learning class distributions, allowing for data augmentation and improved generalization. By focusing on individual class distributions, these models uncover underlying patterns, enabling the generation of synthetic observations that enhance data diversity and bolster generalizability.

\begin{figure}[t]
    \centering
    \includegraphics[width=0.85\linewidth]{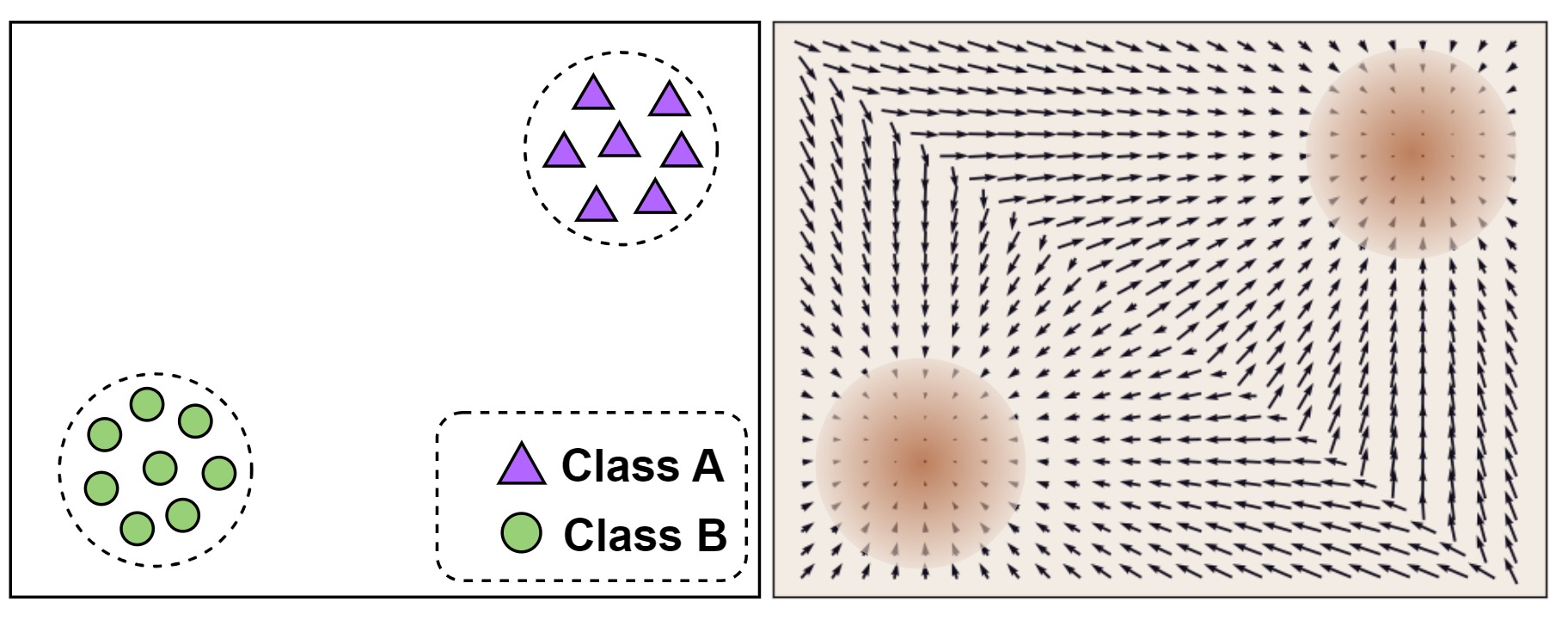}
    \caption{Illustration of score-based approach for (binary) classification task. Class A and B represents two distinct classes in the data distribution space (left), while the score function through denoising score matching represents the direction towards high density regions of respective class (right).}
    \label{fig:MethodologyDiagram}
\end{figure}

Revisiting the classic generative versus discriminative debate, in this paper, we examine how diffusion models (DMs), the current state-of-the-art generative model family, compare against top discriminative models in the context of medical image classification. DM, a recent class of likelihood-based generative models~\cite{ho2020denoising}, have demonstrated remarkable achievements in text-based content creation and editing tasks. Built on sequential noising and denoising methodology, DMs incrementally corrupt initial samples before trying to regenerate them from degraded versions. Training these models with variational inference facilitates effective learning in complicated data manifolds, producing striking results.


Conditional generative models, such as DMs, can be effortlessly transformed into classifiers~\cite{ng2001discriminative}.
Although generative models demonstrated success, often serving as adversarially robust classifiers on elementary datasets like MNIST, this resilience hasn't consistently carried over to more complex datasets, particularly within medical imaging. 
Medical data often shows high similarity and overlapping distributions between classes, making it challenging to delineate clear boundaries. Traditional discriminative models may struggle with such subtle differences. This study addresses these challenges by evaluating score-based diffusion models~\cite{song2020score} as potential alternatives to leading discriminative models for medical image classification, highlighting their capacity to capture underlying distributions in closely related classes and achieve competitive likelihood values. To the best of our knowledge, this is the first study to apply score-based generative models as classifiers for medical images, specifically mammograms, achieving state-of-the-art results for generative classifiers.

\section{Background}

\subsection{Score-Matching}

In machine learning and statistics, it is assumed that data points in a class follow an underlying distribution. Since we rarely know the exact form of this distribution, we estimate it with a model to approximate probabilities. Deep learning can model these complex distributions by learning data patterns, though high-dimensional data remains challenging. Techniques like Generative Adversarial Networks (GANs) address this by modeling the data generation process rather than probability densities, though they can’t yield accurate probability values for individual points.

A better solution is to use Stein scores or score functions, which preserve all the information in the density function~\cite{hyvarinen2005estimation}. 
The score function is the gradient of the logarithmic of the probability density function with respect to the random variable x, $\nabla_\mathbf{x}\text{log }p(\mathbf{x})$, which represents the direction towards the high density data.
Given any probability function, the score can be easily computed, and vice versa, given any score function, we can recover the density function by computing integrals.
Vincent~\cite{vincent2011connection}  introduced denoising score matching (DSM) that allows for faster computation and avoids the computational complexity.
DSM focuses on estimating the score function of a perturbed or noise-contaminated probability distribution instead of the true underlying data distribution.
The score matching objective function can be expressed as the following:
\begin{equation}
    L(\theta) = \frac{1}{2} \mathbb{E}_{q_{\sigma} (\Tilde{\mathbf{x}} | \mathbf{x}) p_d(\mathbf{x})} \left[ \lVert s_m(\Tilde{\mathbf{x}}; \theta) - \nabla_\mathbf{x} \log q_\sigma(\Tilde{\mathbf{x}} | \mathbf{x}) \rVert_2^2 \right]
    \label{eq:DSMgeneral}
\end{equation}
Through the application of the perturbation kernel $q(\cdot)$ with standard deviation $\sigma$, random noise is introduced into the system, to generate a modified perturbed instance $\Tilde{x}$. 

\subsection{Score-based Generative Modeling}

Score-based Generative Modeling with stochastic differential equation (SDE) utilizes the idea of DSM via the simulation of Brownian Motion where the trajectory is influenced by random perturbation.
The concept involves defining a SDE that gradually introduces noise to transition a complex data distribution ($\mathbf{x}$) into a simple prior distribution (isotropic Gaussian, $\mathbf{x}_T$). 
The SDE is defined as~\cite{song2020score},
\begin{equation}
d \mathbf{x} = \mathbf{f}(\mathbf{x}, t) d t + g(t) d \mathbf{w}
\label{eq:generalSDE}
\end{equation}

Here, $\mathbf{f}(\cdot, t): \mathbb{R}^d \to \mathbb{R}^d$ and $g(t) \in \mathbb{R}$ are the drift and diffusion coefficient respectively, alongside standard Brownian motion, $\mathbf{w}$. 
A reverse-time SDE is introduced to reverse the transformation using a time-dependent score function $s_{\theta}(\mathbf{x},t)$, modeled by a neural network with parameters $\theta$. This score function guides the process by directing each time step, progressively removing noise from the prior distribution to recover the original data distribution. To generate new samples, the process starts with random noise $x_T$ and applies the reverse SDE dynamics to derive a sample $x_0$ from the data distribution using below equation~\cite{chen2018neural}:
\begin{equation}
  d\mathbf{x} = [\mathbf{f}(\mathbf{x}, t) - g^2(t)\nabla_{\mathbf{x}}\log p_t(\mathbf{x})] dt + g(t) d\bar{\mathbf{w}},
\end{equation}
In this context, $\nabla_{\mathbf{x}}\log p_t(\mathbf{x})$ is the gradient of the log probability density function, or score function. 
Furthermore, by eliminating the stochastic element, the SDE transforms into a (neural) ordinary differential equation (ODE)~
\cite{chen2018neural}
. 
\begin{equation}
    \text{d}\mathbf{x} = [ \mathbf{f}(\mathbf{x}, t) - \frac{1}{2} g(t)^2\nabla_{\mathbf{x}}\log p_t(\mathbf{x})] \text{d}t
    \label{eq:ODE}
\end{equation}
Utilizing a continuous-time variant of the change of variables formula, it is feasible to calculate the likelihood ($p_0$) of an input image $\mathbf{x}_0$ under the model.

\section{Method}\label{sec:method}


\textbf{Score-based Classifier:}
For any classification task, the main goal is to determine which category or class a new data belongs to. 
One of the approaches involves training individual networks for each classes. 
Each network learns to recognize specific patterns associated with its assigned class.
In this context, we can leverage score-based generative modeling techniques.
Song et al.~\cite{song2020score} proposed the reverse ODE function (Eq.~\ref{eq:ODE}) which can be utilized when the score is known from forward SDE (Eq.~\ref{eq:generalSDE}).
When the score is approximated using any score-based network, i.e. a neural network, the function takes the form of a neural ODE~\cite{chen2018neural}.
By employing neural ODEs, we can compute the density using Eq.~\ref{eq:ODE}.
The final density can be computed using instantaneous change of variables formula, as following:

\begin{equation}
\text{log } p_0((\mathbf{x}(0)) = \text{log } p_T(\mathbf{x}(T)) + \int_0^T \nabla \cdot \tilde{\mathbf{f}}_\theta(\mathbf{x}(t), t) \, dt
\label{eq:unconditionalLogProbability}
\end{equation}

\begin{align}
\text{where, } d\mathbf{x} = \Big\{ \mathbf{f}(\mathbf{x}, t) - \frac{1}{2} \nabla \cdot [\mathbf{G}(\mathbf{x}, t)\mathbf{G}(\mathbf{x}, t)^T] 
& \nonumber \\
- \frac{1}{2} \mathbf{G}(\mathbf{x}, t) \mathbf{G}(\mathbf{x}, t)^T \mathbf{s}_\theta(\mathbf{x},t) \Big\} \, dt &=: \tilde{\mathbf{f}}_\theta(x, t) \, dt
\end{align}

In the above equations, $\mathbf{G}(\cdot, t) : \mathbb{R}^d \rightarrow \mathbb{R}^{d \times d}$ is the diffusion coefficient.
If the data distribution of i.i.d. samples is represented as $p_0 : \mathbf{x}(0) \sim p_0$, $p_T : \mathbf{x}(T) \sim p_T$ denotes the prior distribution with a tractable reverse form, with the noise introduced at time step $T$ ensuring the independence of $p_T$ from $p_0$.
However, when training a classifier for each class separately, the dataset is divided into subsets corresponding to each class. 
Dividing the data into smaller subsets decreases the amount of data available for training each individual model. 
This reduction in data size can lead to overfitting and higher training time. The other approach is training a single network conditioned on class labels, $y \in \mathbb{R}$. 
By conditioning the network on class labels, it learns to identify patterns and features that are specific to each class.
Conditioning a single network on class labels promotes parameter sharing across classes, and simplified model architecture. So, Eq.~\ref{eq:unconditionalLogProbability} can be represented as following;
\begin{equation}
    \text{log } p_0((\mathbf{x}(0) \mid y) = \text{log } p_T(\mathbf{x}(T) \mid y) + \int_0^T \nabla \cdot \tilde{\mathbf{f}}_\theta(\mathbf{x}(t), t, y) dt
    \label{eq:conditionalLogProbability}
\end{equation}

Here, the computation of $\nabla \cdot \tilde{\mathbf{f}}_\theta(x(t), t, y)$ can be expensive for many cases i.e. high dimensional data. 
Grathwohl et al.~\cite{grathwohl2018ffjord} demonstrated efficient computing of the function with Skilling-Hutchinson trace estimator.
We employ this estimator to compute the log-likelihood for any particular class (Eq.~\ref{eq:conditionalLogProbability}).




\textbf{SDE Functions:}
The SDE function can manifest in various forms. At any given continuous time, $t \rightarrow \infty$, it can exhibit either exploding or preserving variance for a sequence positive noise scales, $0 < \beta_1, \beta_2..< 1$. 
These different behaviors are captured by following equations, which are known as variance exploding SDE (VE SDE) and variance preserving SDE (VP SDE) respectively~\cite{song2020score}.

\begin{equation}
   \text{d}\mathbf{x} = \sqrt{\frac{\text{d}[\sigma^2(t)]}{\text{d}t}} \text{d}\mathbf{w}
   \label{eq:ve-sde}
\end{equation}
\begin{equation}
    \text{d}\mathbf{x} = -\frac{1}{2}\beta(t)\mathbf{x}\text{d}t + \sqrt{\beta(t)} \text{d}\mathbf{w}
    \label{eq:vp-sde}
\end{equation}

In addition to the standard VE SDE and VP SDE formulation, it is also possible to derive a modified version of the SDE called the sub-VP SDE.~\cite{song2020score}.
\begin{equation}
    \text{d}\mathbf{x} = -\frac{1}{2}\beta(t)\mathbf{x}\text{d}t + \sqrt{\beta (t) \left(1 - e^{-2\int_{0}^{t} \beta(s) ds}\right)} \text{d}\mathbf{w}
    \label{eq:sub-vp-sde}
\end{equation}



\textbf{Class Likelihood Computation:}
In general, when employing a conditional generative model for classification, Bayes' theorem can be applied to the model's predictions $p_\theta(\mathbf{x} \vert y_i)$ and the prior $p(y)$ over labels $y_i$ where $i \in \{1, 2,...,n\}$. For any uniform prior assumption ($p(y_i) = \frac{1}{n}$), the Bayes' equation is given by:

\begin{align}
    p_\theta\left( y_{i} \mid \mathbf{x}\right) &= \frac{p\left( y_{i}\right) p_{\theta }\left( \mathbf{x} \mid y_{i}\right)}{\sum\limits_{j=1}^n p\left( y_{j}\right) p_{\theta }\left( \mathbf{x} \mid y_{j}\right)} = \frac{p_{\theta }\left( \mathbf{x} \mid y_{i}\right)}{\sum\limits_{j=1}^n p_{\theta }\left( \mathbf{x} \mid y_{j}\right)}; \notag \\
    &\text{where} \;\; i, j \in \{1, 2, \ldots, n\}. \label{eq:bayes}
\end{align}


\begin{figure}[t]
  \begin{minipage}{0.45\textwidth}
    \begin{algorithm}[H]
      \caption{Training Algorithm}
      \label{alg:training}
      \begin{algorithmic}
        \STATE {\bfseries Input:} input $\mathbf{x}(0) \in \mathbb{R}^{C \times H \times W}$, class label $y \in \mathbb{R}$, SDE model $\phi$
        \STATE Initialize score model parameters $\mathbf{s}_\theta$
        \REPEAT
        \STATE $z_i \sim \mathcal{N}(0, \sigma^2\textbf{I})$; $t_i \sim \mathcal{N}(0, T)$
        \STATE $\mu$, $\sigma$ $\leftarrow \phi(\mathbf{x}_i(0))$
        \STATE $\mathbf{x}_i(t) \leftarrow \mu + \sigma * z_i$
        \STATE $\text{Score} \leftarrow \mathbf{s}_\theta(\mathbf{x}_i(t), y_i, t_i)$
        \STATE $L_{\text{DSM}} \leftarrow \text{Loss} (\text{Score})$ using Eq.~\ref{eq:DSMloss}
        \UNTIL{Convergence}
      \end{algorithmic}
    \end{algorithm}
  \end{minipage}%
  \hfill
  \begin{minipage}{0.50\textwidth}
    \begin{algorithm}[H]
      \caption{Inference Algorithm}
      \label{alg:testing}
      \begin{algorithmic}
        \STATE {\bfseries Input:} input $\mathbf{x}(0) \in \mathbb{R}^{C \times H \times W}$, label $y_{\text{gt}} \in \mathbb{R}$, SDE model $\phi$, score model $\mathbf{s}_\theta$
        \FOR{$j$ = {1, 2, .. n}}
        \STATE $z_i \sim \mathcal{N}(0, \sigma^2\textbf{I})$; $t_i \sim \mathcal{N}(0, T)$
        \STATE $\mu$, $\sigma$ $\leftarrow \phi(\mathbf{x}(0))$
        \STATE $\mathbf{x}(t) \leftarrow \mu + \sigma * z_i$
        \STATE $\text{Score} \leftarrow \mathbf{s}_\theta(\mathbf{x}(t), y_j, t_i)$
        \STATE $\text{Compute }p_0(y_j \mid (\mathbf{x}(0))$ using Eq.~\ref{eq:conditionalLogProbability}
        \ENDFOR
        \RETURN $p_0(y_\text{gt} \mid (\mathbf{x}(0))$ using Eq.~\ref{eq:bayes}
      \end{algorithmic}
    \end{algorithm}
  \end{minipage}
\end{figure}

\textbf{Training Objective:}
We applied a time-dependent conditional score model, $\mathbf{s}_\theta(\mathbf{x}(t), t, y)$, to train using a weighted sum of reformulated conditional denoising score matching,~\cite{song2020score}

\begin{multline}
    \arg \min_\theta \mathbb{E}_{t\sim \mathcal{U}(0, T)} [\lambda(t) \mathbb{E}_{\mathbf{x}(0) \sim p_0(\mathbf{x})}\mathbb{E}_{\mathbf{x}(t) \mid \mathbf{x}(0) \sim p_{0t}(\mathbf{x}(t) \mid \mathbf{x}(0))} \\ 
    [ \|\mathbf{s}_\theta(\mathbf{x}(t), t, y) - \nabla_{\mathbf{x}(t)}\log p_{0t}(\mathbf{x}(t) \mid \mathbf{x}(0), y)\|_2^2]]
    \label{eq:SGMobj}
\end{multline}

If the perturbation kernel utilizes a Gaussian distribution, the DSM objective (Eq.~\ref{eq:DSMgeneral},~\ref{eq:SGMobj}) can be reformed as the following using empirical means,~\cite{vincent2011connection},

\begin{equation}
    L_{\text{DSM}}(\theta) = \frac{1}{2N} \sum_{i=1}^{N} \left\lVert \mathbf{s}_{\theta}(\mathbf{x}(t), t, y) - \frac{\mathbf{x}(0) - \mathbf{x}(t)}{\sigma^2} \right\rVert_2^2
    \label{eq:DSMloss}
\end{equation}


\section{Experimental Details}

$\;\;\;$\textbf{Dataset:} 
The CBIS-DDSM dataset~\cite{lee2017curated} contains 1,231 images (629 benign, 602 malignant), with a test set of 361 images (216 benign, 145 malignant). The INbreast dataset~\cite{moreira2012inbreast} includes 106 images with breast masses; we used 60 images (25 benign, 35 malignant) for training, leaving 46 images (10 benign, 36 malignant) for testing. The VinDr-Mammo dataset~\cite{Nguyen2022.03.07.22272009} comprises 20,000 mammograms from 5,000 women, split into 16,000 for training (15,210 benign, 790 malignant) and 4,000 for testing (3,802 benign, 198 malignant). Since VinDr provides only BIRADS classifications, so we categorized BIRADS values 1–3 as benign and 4–6 as malignant.

\textbf{Implementation Details:} For our implementation, we decided to select $t$ to be random with a range between $\beta_{min}=0.1$ and $\beta_{max}=20$ for VP SDE and sub-VP SDE, and $\sigma_{min}=0.01$ and $\sigma_{max}=50$ for VE SDE. 
All random variable utilized in our experiments follows a Gaussian distribution.
As part of our experimental setup, we implemented a conditional UNet~\cite{ronneberger2015u} architecture as the backbone for our score model.
All models used a batch size of 32 and the Adam optimizer with an initial learning rate of $10^{-4}$.
We have also employed early stopping and Exponential Learning Rate scheduler with a gamma value of 0.25.
Additionally, to circumvent the challenge of non-differentiability, we have chosen the time range as $t \in [\epsilon, 1]$ where $\epsilon=10^{-5}$. 


\begin{table}[!t]
  \centering
  \newcommand{\hlc}{\textbf}
  \caption{Evaluation of various architectural configurations in the proposed approach across three datasets. Here, Acc.: Accuracy, AUC: Area under the curve}
  \label{table1}
  \scalebox{.85}{
    \begin{tabular}{ccccccc}
      \hline
      {\multirow{2}{*}{\shortstack{SDE \\ Function}}} & \multicolumn{2}{c}{CBIS} & \multicolumn{2}{c}{INbreast} &
      \multicolumn{2}{c}{Vin-Dr} \\
      \cmidrule(lr){2-3}
      \cmidrule(lr){4-5}
      \cmidrule(lr){6-7}
      &  Acc. & AUC & Acc. & AUC & Acc. & AUC \\
      \hline
      VPSDE & \textbf{63.65} & \textbf{71.75} & \textbf{75.00} & \textbf{78.85} & 84.78 & \textbf{64.16} \\
      VESDE & 62.60 & 54.01 & 63.64 & 42.85 & 84.92 & 48.77 \\
      SubVPSDE & 58.73 & 51.69 & 63.64 & 36.5 & \textbf{85.72} & 50.78 \\
      \hline
    \end{tabular}
  }
\end{table}


\section{Results and Discussion}

To illustrate the strength and robustness of our model, we performed experiments using a variety of datasets, each presenting distinct limitations. The CBIS-DDSM dataset comprises limited yet balanced data, while VinDr offers significant imbalance, and INbreast contains highly limited data. Additionally, CBIS-DDSM mammograms are scanned, leading to lower image quality, whereas the images in the INbreast and VinDr datasets are digitally enhanced, resulting in higher quality. Crucially, we utilized whole mammographic images for all datasets to preserve consistency with clinical practice, where radiologists analyze entire mammograms rather than individual patches.
In Table~\ref{table1}, we compare performances across different architectural settings, reporting essential metrics. To select the optimal architecture, we explored three distinct versions of our model: VP SDE, VE SDE, and sub-VP SDE.
Given that VPSDE consistently performed well across datasets, we selected it as our proposed architecture.


To fortify claims about the robustness of generative classifiers compared to discriminative classifiers, Table 2 presents a thorough comparative analysis featuring four popular discriminative classifiers: ResNet50~\cite{he2016deep}, Inception V3~\cite{szegedy2016rethinking}, Vision Transformer~\cite{dosovitskiy2020image} and Swin Transformer~\cite{liu2021swin}. 
For the discriminative models, we maintained the same experimental regime by keeping all the settings static for fair comparison. Importantly, we documented accuracy, specificity, sensitivity and AUC (Table~\ref{table2}). 
As VinDr and INbreast are highly imbalanced datasets, AUC is significantly impacted, as shown in~\cite{fernandez2018learning}.

We contend that discriminative models perform optimally when dealing with clearly distinguishable data distributions across classes. However, this assumption does not always hold true in medical imaging, as the class distributions tend to be closely related or even overlapping. To illustrate this concept, in the case of mammogram data; despite all images inherently representing breast images, each contains masses that may be either benign or malignant. As a result, the data distribution for malignant and benign classes exhibits substantial similarity and might overlap. Discriminative models encounter challenges in delineating a boundary, often succumbing to overfitting and incorrectly classifying most instances as a particular class, ultimately yielding extremely low or negligible sensitivity and specificity. 
In contrast, our generative classifier is trained to learn the underlying data distribution pertinent to each class. 
During the inference phase, it makes predictions based on this learned distribution, determining whether the input belongs to Class A or Class B (see Fig.~\ref{fig:MethodologyDiagram}). Consequently, our generative classifier demonstrates the capability to classify both malignant and benign instances, though poorly, thereby minimizing false positives and negatives with minimal training and a vanilla conditional UNet.

\begin{table}[!t]
  \centering
  \newcommand{\hlc}{\textbf}
  \caption{Comparative assessment of the proposed architecture with state-of-the-art discriminative models. Here, Acc.: Accuracy, Spe.: Specificity Sen.: Sensitivity}
  \label{table2}
  \scalebox{0.75}
  {
  \begin{tabular}{*{7}{c}}
    \hline
    \multirow{2}{*}{\shortstack{Dataset}} & \multirow{2}{*}{\shortstack{Metrics}} & \multicolumn{5}{c}{Model}
    \\
    \cmidrule{3-7}
    & & ResNet50 & InceptionV3 & VIT & Swin-T & Ours
    \\
    \hline
    \multirow{4}{*}{\shortstack{CBIS}} & Acc. & 54.35 & 59.54 & 58.20 & 55.96 & \textbf{63.65} \\
    & Spe. & \textbf{80.00} & 58.10 & 58.40 & 51.39 & 72.83  \\
    & Sen. & 47.22 & 41.50 & 44.10 & 41.38 & \textbf{58.21} \\
    & AUC & 61.38 & 47.78 & 52.44 & 52.41 & \textbf{71.75} \\
    \hline
    \multirow{4}{*}{\shortstack{INbreast}} & Acc. & 45.65 & 66.52& 56.41 & 68.56 & \textbf{75.00} \\
     & Spe. & 37.60 & 30.27 & 37.10 & 20.00  & \textbf{44.44} \\
    & Sen. & 65.30 & 70.50 & 66.70 & 78.60 & \textbf{82.86} \\
    & AUC & 50.83 & 60.78 & 51.87 & 66.66 & \textbf{78.85} \\
    \hline
    \multirow{4}{*}{\shortstack{VinDr- \\ Mammo}} &  Acc. & 52.20& 75.96 & 64.01 & 84.40 & \textbf{87.78}\\
    & Spe. & 51.95 & \textbf{92.24} & 49.81 & 87.87 & 84.78\\
    & Sen. & \textbf{57.07} & 37.77 & 52.34 & 17.68 & 44.14\\
    & AUC & 57.59 & 58.29 & 51.01 & 53.67 & \textbf{64.16}\\
    
    \hline
  \end{tabular}
  }
\end{table}

\section{Conclusion}


In this study, we highlight the efficacy of adopting score-based generative classifiers for managing medical datasets marked by limited data and skewed class distributions. Our results underscore the merits of leveraging score-based generative models for classification tasks, surpassing several discriminative models' performance. In contrast to discriminative models, which are susceptible to overfitting, our approach adeptly captures underlying patterns, thereby demonstrating robust performance even with limited data.  
In future research, we aim to extend this theoretical perspective towards segmentation task in medical images.

\section{Acknowledgments}
\label{sec:acknowledgments}
This work has been supported in part by the Nevada DRIVE program   
(Doctoral Research in Innovation, Vision and Excellence), 
the National Institute of General Medical Sciences of the National Institutes of Health under Grant No. P30GM145646 and the National Science Foundation OIA-2148788  under Award No. 2201599.

\section{Compliance with Ethical Standards}
This research study was conducted retrospectively using human subject data made available in open access by \cite{lee2017curated, moreira2012inbreast, Nguyen2022.03.07.22272009}. 
Ethical approval was not required as confirmed by the license attached with the open access data.

\bibliographystyle{IEEEbib}
\bibliography{strings,refs}

\end{document}